\documentclass{iosart2c}

\usepackage[T1]{fontenc}
\usepackage{times}%

\usepackage{natbib}
\setcitestyle{numbers,square}
\usepackage{amsmath}
\usepackage{dcolumn}
\usepackage{graphics}
\usepackage{graphicx}%
\usepackage{multirow}%
\usepackage{amsmath,amssymb,amsfonts}%
\usepackage{amsthm}%
\usepackage{mathrsfs}%
\usepackage[title]{appendix}%
\usepackage{xcolor}%
\usepackage{textcomp}%
\usepackage{manyfoot}%
\usepackage{booktabs}%
\usepackage{algorithm}%
\usepackage{algorithmicx}%
\usepackage{algpseudocode}%
\usepackage{listings}%
\usepackage{array}%
\usepackage{natbib}
\usepackage[T1]{fontenc}
\usepackage{times}%

\newcolumntype{d}[1]{D{.}{.}{#1}}

\firstpage{1} \lastpage{5} \volume{1} \pubyear{2009}

\begin{document}
\begin{frontmatter}                           

%
\title{Emotion-cause pair extraction method based on multi-module interaction}

\runningtitle{Instructions for the preparation of a camera-ready paper in \LaTeX}

\author[A,B]{\fnms{Mingrui} \snm{Fu} \thanks{Corresponding author. E-mail: a2813176983@126.com.}},
\author[A,B]{\fnms{Weijiang} \snm{Li}} and 
\author[A,B]{\fnms{Zhengtao} \snm{Yu}}

\address[A]{Faculty of Information Engineering and Automation, Kunming University of Science and Technology, Yunnan,Kumming 650500, P.R.china}
\address[B]{Yunnan Key Laboratory of Artificial Intelligence, Kunming University of Science and Technology, Yunnan,Kumming 650500, P.R.china}

\begin{abstract}
The purpose of emotion-cause pair extraction is to extract the pair of emotion clauses and cause clauses. On the one hand, the existing methods do not take fully into account the relationship between the emotion extraction of two auxiliary tasks. On the other hand, the existing two-stage model has the problem of error propagation. In addition, existing models do not adequately address the emotion and cause-induced locational imbalance of samples. To solve these problems, an end-to-end multitasking model (MM-ECPE) based on shared interaction between GRU, knowledge graph and transformer modules is proposed. Furthermore, based on MM-ECPE, in order to use the encoder layer to better solve the problem of imbalanced distribution of clause distances between clauses and emotion clauses, we propose a novel encoding based on BERT, sentiment lexicon, and position-aware interaction module layer of emotion motif pair retrieval model (MM-ECPE(BERT)). The model first fully models the interaction between different tasks through the multi-level sharing module, and mines the shared information between emotion-cause pair extraction and the emotion extraction and cause extraction. Second, to solve the imbalanced distribution of emotion clauses and cause clauses problem, suitable labels are screened out according to the knowledge graph path length and task-specific features are constructed so that the model can focus on extracting pairs with corresponding emotion-cause relationships. Experimental results on the ECPE benchmark dataset show that the proposed model achieves good performance, especially on position-imbalanced samples.
\end{abstract}


\end{frontmatter}

\section{Introduction}
 Emotion-cause Analysis (ECA) aims to analyze the hidden emotion-causes in texts, and it plays an important role in public opinion analysis, online comments, etc. As a part of natural language, emotion categories include happiness, sadness, anger, fear, etc. Identifying various emotions in text and make machines identify the causes of emotions has become one of the hot topics in natural language processing. Emotion-cause analysis includes three sub-tasks: emotion extraction (EE), emotion-cause extraction (ECE) and emotion-cause pair extraction (ECPE). As a new task of emotion cause analysis, emotion cause pair extraction requires both a comprehensive understanding of the document and detailed analysis of specific aspects.

\begin{figure}[ht]
\centering
\includegraphics[height=3.5cm,width=7.6cm]{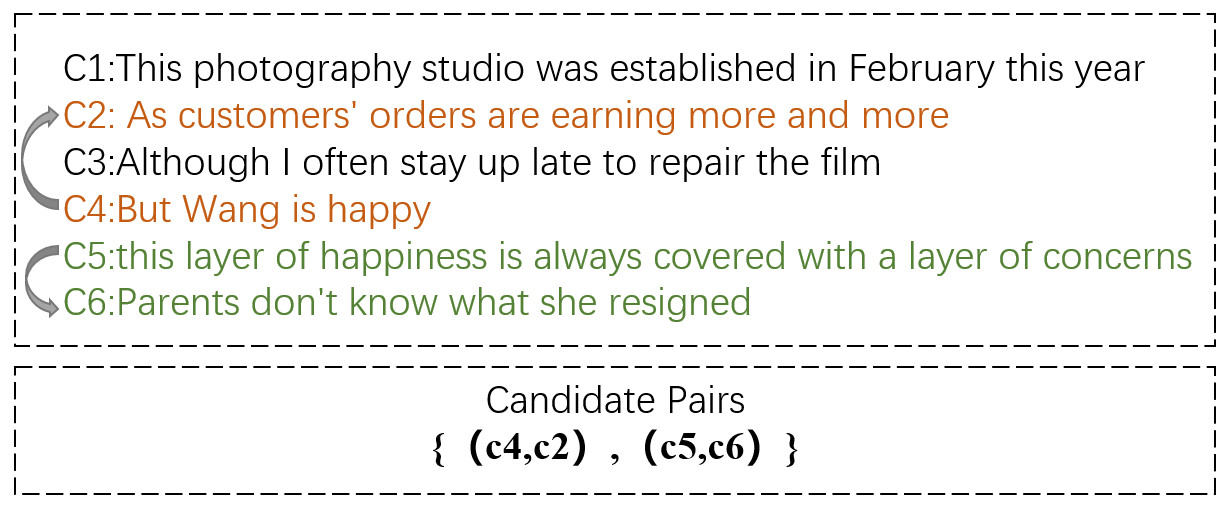}
\caption{Extraction examples of emotion-cause pairs.}
\label{Figure1}
\end{figure}

  The research on emotion-cause extraction proposed by Lee et al.\cite{1_Lee2010text-driven} needs to manually annotate emotions and then extract the corresponding causes, which not only limits its application in real task scenarios, but also ignores the correlation between emotions and causes. Identifying the underlying causes for certain expressions in text is usually regarded as a clause classification task. Xia et al.\cite{2_Xia2019semi} proposed a new task of extracting Emotion-cause pairs is proposed to solve the problems caused by the Emotion-cause extraction task. emotion-cause pair extraction aims to extract both emotion clauses and their underlying causes from documents. For example, Fig.\ref{Figure1} gives an example of Emotion-cause pair extraction.

  There are six clauses in Clause c4 contain the emotion "happy " and its cause clause is c2. Clause c5 contains the emotion " concerns ", and its corresponding cause clause is c6. The final result of emotion-cause pair extraction is: {(c4, c2), (c5, c6)}.
  
  The two-stage framework proposed by Xia et al. \cite{2_Xia2019semi} In the first step, the emotion clause and the cause clause are extracted respectively through the model. In the second step, the model performs the Cartesian product of the emotion clause and the cause clause to obtain all the emotion-cause pairs, and then filters the candidate clause pairs through the logistic regression model to extract the correct emotion-cause pairs. However, the relationship between main task emotion extraction and two auxiliary tasks emotion extraction and cause extraction is complicated and the three tasks indicate each other. Previous studies have ignored the problem, or unilaterally considered the contribution of emotion prediction to cause extraction, rather than the interaction.

  \begin{figure}[ht]
\centering
\includegraphics[height=3.5cm,width=7cm]{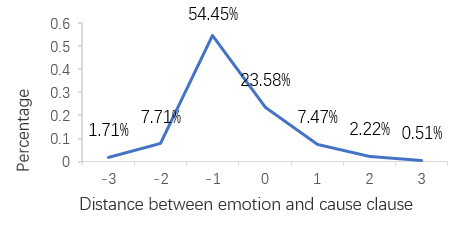}
\caption{Relative position of emotion-causes.}
\label{Figure2}
\end{figure}
 First of all, the existing models filter out emotion-cause pairs with a certain distance by setting a threshold for the distance between the emotion clause and the cause clause. The Chinese emotion cause corpus proposed by Xia et al. \cite{2_Xia2019semi}. The relative position distribution of its emotion clauses and corresponding cause clauses is shown in the figure \ref{Figure2}. Although nearly 95.8\% of emotion-cause pairs have a distance of less than 3 from the emotion clause to the cause clause, considering only the relative position information and ignoring the semantic dependence increases the problem caused by the unbalanced distribution of the relative distance between emotion and cause.

  Secondly, on the one hand, although the existing models notice that the emotion clause and the cause clause are interrelated, they do not make full use of the correlation information between the two, nor do they take into account the impact of emotion words and auxiliary task emotion prediction on the main task . With the help of emotion-cause pairs, a bidirectional association of emotion and cause extraction is constructed. And the performance of the two auxiliary tasks may improve if done interactively rather than independently. On the other hand, in the language expression, the emotion and the cause clause in the effective emotion-cause pair have the same emotion consistency. There is also an intricate many-to-many mapping between emotion and cause. To alleviate the positional bias problem, we use commonsense knowledge to enhance the semantic dependency between candidate clauses and emotion clauses.

  A multi-task joint model including a document encoding module, a subtask prediction module and an emotion-cause pair prediction module. First, inter-clause interaction representation encoding is performed by a multi-level shared module. Secondly, considering that the semantics between clauses are helpful to find contextual causality, the emotion and cause are predicted respectively through the subtask module, and added to the coding of the other side. Then, the emotion-cause pair prediction module screens out suitable samples of position imbalance based on the knowledge graph, and then encodes the emotion-cause pair through Transformer \cite{3_Vaswani2017Attention}, and then per-forms location modeling and extraction. We evaluate our method on the ECPE benchmark dataset.

  In addition, in order to use the encoding layer to alleviate the impact of sample Imbalanced distribution of emotion clauses and cause clauses, We replace the original encoding module with a position-aware encoding module and propose the MM-ECPE(BERT) model. The model fully considers the interaction and consistency among the three tasks . Specifically, the sentiment lexicon knowledge is first introduced into the sentence encoding layer to obtain sentence encoding enhanced by external knowledge. Then introduce the relative position deviation item into the PAIM module to capture the position deviation between emotion words and cause words, and then perform position modeling and extraction. Next, inter-clause interaction representations are encoded through a shared encoding layer.  Second, considering that the consistency between emotion and cause clauses helps to find contextual causality, the subtask module predicts emotion and cause separately, and performs inter-task interactions. We evaluate the effectiveness of our method on the ECPE benchmark dataset. 

  Experimental results show that the method outperforms existing methods. Further analysis proves the effectiveness of the proposed model. In summary, the main contributions of this article are as follows:
  \begin{itemize}
\item We propose a joint learning model that includes emotion extraction, cause extraction, and emotion-cause pair extraction. Through the shared module of emotion extraction and emotion cause extraction tasks, the information between the three tasks can be better utilized.
\item Introduce the knowledge graph into the construction of imbalanced samples to alleviate the problem of imbalanced label distribution. 
\item Based on the MM-ECPE model, using a position-aware interactive encoding layer, the MM-ECPE (BERT) model is proposed to better solve the problem of imbalanced distribution of emotion and cause clauses.
\item Experimental results show that the performance of the proposed method on the raw ECPE dataset is comparable to the existing state-of-the-art methods.
\end{itemize}

\section{Related work}
\subsection{Emotion cause extraction}

Emotion-cause Extraction (ECE) is based on statistical learning methods, machine learning methods and deep learning methods. In the early days, Gui et al. \cite{4_Gui2017question} used multi-core SVM classifiers to construct datasets, and some scholars completed ECE tasks based on the public and available datasets they constructed. Experiments show that compared with traditional classification algorithms, support vector machines have better performance and can obtain better classification results in a lower training time. After that, this dataset serves as the benchmark dataset for ECE. On this basis, the researchers extended it to other types of applications, such as speech recognition, face recognition, and natural language processing. With the deepening of deep representation learning research, the attention mechanism is widely used. For the ECE task, a set of task processing methods based on deep learning \cite{4_Gui2017question,5_Li2018co-attention,6_Xinyi2019Multiple} was proposed. These methods attempt to integrate text sequence information and the relationship between emotional words and clauses, in order to improve the accuracy of extraction.For example, Li et al. \cite{5_Li2018co-attention} proposed that the context of emotion helps to find valuable clues for corresponding causes, and the joint attention module is designed to exploit the context of emotions in ECE tasks. Yu et al. \cite{6_Xinyi2019Multiple} noticed the positive effect of the relationship between clauses on the task, thus proposing a hierarchical framework. The framework considers not only the emotion description, but also the interaction between clauses' correlation and semantic information between clauses. In addition, Ding et al. \cite{7_Li2018co-attention} found that emotion description, label information and relative position between clauses are of great significance to the extraction of emotion causes. In addition, some researchers have also focused on how to obtain more relevant features from text data, such as part of speech, syntactic structure, and clause length, etc. Xia et al. \cite{8_Ding2019RTHN} used Transformer to model the relationship between clauses and extracted Emotion-causes. These methods are all based on the bag-of-words model. Hu et al.\cite{9_Hu2014FSS-GCN}  used graph convolutional networks to encode the semantic and structural information of clauses. ECE also has two disadvantages: on the one hand, ECE must first annotate emotions before extracting causes, which greatly limits its use in real-world scenarios; On the other hand, the practice of first annotating emotions and then extracting reasons ignores the fact that these emotions explain each other. Before the Emotion-causes are extracted, the emotions must be manually marked. For this cause, Xia et al. \cite{2_Xia2019semi} proposed the ECPE task in the article.

\subsection{Emotion cause pair extraction}
\subsubsection{Two-stage methods}
The first stage of the two-stage model regards cause extraction and emotion extraction as two independent tasks, and completes two subtasks respectively. The second stage pairs emotion clauses and cause clauses through Cartesian product, and then screens out emotion-cause pairs with relation through filters. The previous work\cite{2_Xia2019semi}  was the first to address the emotion-cause pair extraction task and proposed a two-stage model. Specifically, the method first extracts emotion clauses and cause clauses separately, and then trains a classifier to filter negative pairs. Although two-stage solutions have been shown to be effective, the system of this two-stage model is not optimal for ECPE as it faces error propagation. In addition, the clause extraction model and the pair extraction model are independent, with different feature spaces and output label spaces, resulting in the inability to fully exploit the mutual benefit of related tasks.
\subsubsection{End-to-end methods}
 Scholars have done a lot of work to solve the ECPE task in an end-to-end method. \textbf{} For example, Wei et al. \cite{10_Wei2020Effective} used the relationship between clauses. They rank the opinions, and select the highest-ranked pairs as emotion-cause pairs . Ding et al. \cite{11_Ding2020ECPE-2D} used a 2D Transformer to simulate the interaction between emotion-cause pairs. They propose a new method to identify whether there is an association or a causal relationship by analyzing the difference in frequency of occurrence at different positions between two clauses. They integrate emotion-cause pair interaction and emotion-cause pair prediction in a unified framework, and completed the extraction of emotion-cause pair. In addition, Ding et al. \cite{12_Ding2020End-to-end} believe that only when emotions are identified can we better understand the causes. To address this issue, they proposed two different dual frames. The first framework EMLL represents emotion clauses in the document, and then extracts the corresponding cause clause from the context of the emotion clause through multi-label learning. This method can automatically obtain the semantic content and emotion characteristics of clauses from text in-formation. The second framework CMLL takes each clause in the document as a reason clause, and then extracts the corresponding sentiment clause from the context of the reason clause through multi-label learning. Since the two methods are based on clause analysis, they can also be regarded as the extension and development of text classification technology. Yuan et al. \cite{13_Yuan2020Emotion} designed a new labeling scheme and proposed a sequence labeling model based on BiLSTM for extracting emotion-cause pairs. Tang et al. \cite{14_Tang2019Joint}  propose a joint learning framework for EE and ECPE. Wu et al. \cite{15_Wu2020multi} proposed a joint learning model to jointly predict emotions, causes and emotion-cause pairs, and explored the relationship between three tasks. Song et al.\cite{16_Song2021An} regarded the ECPE task as the bidirectional connection between emotion and cause. Yu et al. \cite{17_Yu2021Mutually} proposed a supportive joint model, adding two sub-tasks to the model to construct the interaction between emotion extraction and cause extraction. Chen et al. \cite{18_Tang2019Joint} jointly extract emotions, causes, and emotion-cause pairs through joint learning, and specifically model the interactions between tasks. The framework adopts a recurrent synchronization mechanism to allow appropriate utilization of correlations between different tasks to iteratively improve its predictions. Wu et al. \cite{19_Wu2023Pairwise} proposed a new tagging framework, PTF, to solve the ECPE task in an end-to-end method. PTF converts emotion extraction, cause extraction, and chance detection into emotion and cause into relational classification of clauses, thus successfully integrating the whole ECPE into a globally optimized and unified labeling task. Chen et al. \cite{20_Chen2022Recurrent} proposed a new ECPE framework called RSN. It jointly extracts emotions, causes, and emotion-cause pairs through joint learning, and models the interactions between different tasks. The recursive synchronization mechanism in our framework is able to correctly exploit the correlations between different tasks to iteratively improve their predictions. Huang et al.\cite{21_Huang2022Deep} proposed a deep neural network model based on span association prediction for emotion pause pair extraction. Using the span notation, we approach this task from a grammatical idiomatic point of view using the notion of a span. At the same time, the span association matching method is adopted to strengthen the research on the deep meaning of the sentence, pair the sentences in the document, and accurately obtain candidate emotion-cause pairs.

 In summary, the previous end-to-end methods achieved good results but fail to alleviate the sample location imbalance problem well. Therefore, we need to further study how to increase the diversity of labels to improve model performance. To alleviate the aforementioned issues, we propose an end-to-end multi task learning model MM-ECPE, which uses a strategy based on constructing positional deviation samples to construct a training set. The MM-ECPE (BERT) model adopts the position awareness module (PAIM) to capture the position correlation between emotion and cause to solve the position bias problem. Unlike previous methods that implicitly capture partial reciprocity, our model explicitly captures all the interaction information between the aforementioned emotion and cause clauses.

\begin{figure*}[ht]
\centering
\includegraphics[height=10cm,width=12cm]{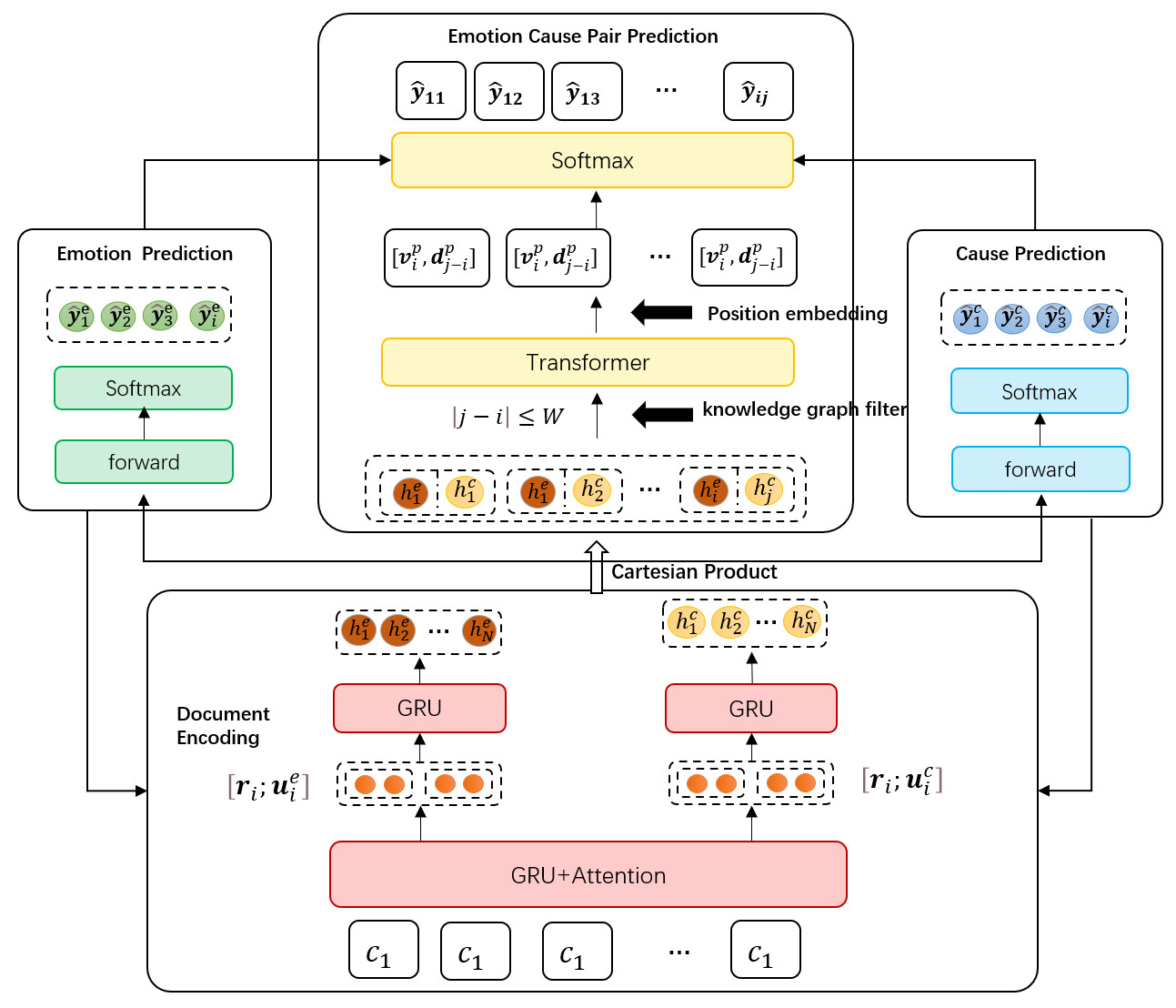}
\caption{MM-ECPE model.}
\label{Figure3}
\end{figure*}

\section{Model}

Given a document $ D=\left\{c_1,c_2,...,c_i\right\}_{i=1}^M $ , the number of clauses contained in it is M. Clauses are expressed as $ c_i=\left\{w_{i1},w_{i2},...,w_{ij}\right\}_{j=1}^N $ , where N is the length of the clause. The emotion cause pair composed of emotion clause  $ <{c_e,c}_c> $ and cause clause $ c_cis $  extracted through the model $ c_e $ . Finally, predict the probability distribution of emotion-causal pairs if they constitute a causal relationship $ {c_e,c}_c $ .

  The overall architecture of the proposed model is shown in Fig.\ref{Figure3}. First, consider that the relative distance distributions of emotions and causes are wildly imbalanced. We decide to leverage commonsense knowledge to enhance the semantic dependencies between candidate cause clauses and emotion clauses to alleviate the positional bias problem. The multi-level sharing module is to encode clauses and learn the interrelationships between clauses. Among them, the clause encoding captures the representation between words, the context encoding between words in clauses, and the clause information encoding for learning the relationship between words. For the previous tasks, the relationship between emotion prediction and cause prediction is ignored , and the two auxiliary tasks of emotion prediction and cause prediction are considered, and the two tasks are interacted. After multiple rounds of iterative prediction of emotion clauses and cause clauses, two prediction modules generate predictions based on clause representations and information provided by other modules. In addition, in order to alleviate the problem of position deviation, We use the emotion clauses and cause clauses extracted from the document to filter the Imbalanced distribution of emotion clauses and cause clauses labels through the knowledge graph path hops combined with the sorting algorithm to establish a new test set, which is more conducive to the use of Semantics in Knowledge Graphs.

  The overall framework of the model consists of multiple modules: multi-level sharing module, sub-task prediction module, and Emotion-cause pair prediction module. This article presents the details of each layer in the following sections.

  \subsection{Multi-level shared modules}\label{subsec2}
  \subsubsection{Clause Encoding}\label{subsubsec2}
  
  To capture specific features of emotion and cause, for a given document, the contextual encoding module use attention and GRU to capture important contextual features and generate clause representations. We first use a bidirectional GRU (Gated Recurrent Neural Network) \cite{22_Cho2014Learning} to build contextual information. GRU is a variant of LSTM. It has fewer parameters than LSTM, so the training speed is faster.

  Firstly, each clause in the document is $ \left\{w_{i j}\right\}_{j=1}^N $ mapped to a sequence of vectors $ \left\{{v}_{ij}\right\}_{j=1}^N $ through the embedding matrix $ {E}_w\in R^{d_w\times V}$. where $ d_wis $ the dimension of the word embedding, $ Vis $ the vocabulary size. Consequently, word-level contextual information is encoded using a bidirectional gated recurrent neural network (GRU):
  
  \begin{eqnarray}
  \boldsymbol{h}_{i j}=\left[\overrightarrow{\operatorname{GRU}_{w}}\left(\boldsymbol{v}_{i j}\right) ; \overleftarrow{\operatorname{GRU}_{w}}\left(\boldsymbol{v}_{i j}\right)\right]
  \label{g1}
  \end{eqnarray}

  where $ {h}_{ij}\in R^{2d_h} $ is the hidden state of the jth word in the ith clause and $ d_his $ the number of hidden units in the GRU.

  Consider that not all words in a clause are equally informative in expressing emotion or cause. We use an attention layer to enable the model to focus on more important words. The model is able to focus on more informative words. Among them, the attention weight of the j-th word in the i-th clause is calculated as follows:
  \begin{eqnarray}
  \boldsymbol{a}_{i j}=\boldsymbol{w}_{a}^{\top} \tanh \left(\boldsymbol{W}_{a} \boldsymbol{h}_{i j}+\boldsymbol{b}_{a}\right)
  \end{eqnarray}
  \begin{eqnarray}
  \boldsymbol{a}_{i j}=\frac{\exp \left(\boldsymbol{a}_{i j}\right)}{\sum_{j=1}^{M} \exp \left(\boldsymbol{a}_{i j}\right)}
  \end{eqnarray}

  where $ {W}_a\in R^{2d_h\times2d_h} $ , $ {b}_a\in R^{2d_h}$ and ${w}_a\in R^{2d_h} $ are learnable parameters.
 
  the i-th clause is as follows:
  \begin{eqnarray}
  {r}_{i}=\sum_{j=1}^{M} 
  {\alpha}_{i j} 
  {h}_{i j}  
  \label{g4}
  \end{eqnarray}

 \subsubsection{Learning the relationship between clauses}\label{subsubsec2}

 Inter-clause encoding aims to encode the words within a clause and the contextual information between clauses. Among them, the two modules of emotion clause coding and cause clause coding are similar in structure, and We take emotion clause coding as an example.

 Since the information of cause clauses and emotion cause pairs can help the model to better recognize emotion clauses. In order to exploit the correlation among several modules, the predictive labels of cause clauses and emotion-cause pairs are embedded into the emotion information to fully utilize the correlation among the three.

 The input of emotion clause encoding consists of the following three parts: the clause representation from the context encoding module $\left\{\boldsymbol{r}_{i}\right\}_{i=1}^{N}$, the cause label embedding from the cause prediction task $\left\{\boldsymbol{u}_{i}^{c}\right\}_{i=1}^{N}$, and the two parts are concatenated and input to a bidirectional GRU network to encode specific emotions at the clause level Information below.
 
 \begin{eqnarray}
  t_{i}^{e}=\left[r_{i} ; u_{i}^{c}\right] \label{g5}
  \end{eqnarray}
  \begin{eqnarray}
 h_{i}^{e}=\left[\overrightarrow{\operatorname{GRU}_{e}}\left(t_{i}^{e}\right) ; \overleftrightarrow{\operatorname{GRU}_{e}}\left(t_{i}^{e}\right)\right] \label{g6}
  \end{eqnarray}

  where $ h_i^e\in R^{2d_h} $ is the clause representation of the ith clause.

  The predictions are mapped to a sequence of vectors $ \left\{u_i^e\right\}_{i=1}^Nvia $ the embedding matrix $ E_e\in R^{d_l\times2} $ , where $ d_lis $ the dimensionality of the label embeddings. Embed the updated emotion label embedding information into other prediction modules.

  \subsection{Subtask prediction module}\label{subsec2}

  The purpose of emotion and cause prediction task is to learn whether a clause is a emotion clause or a cause clause. Next, the clause representation of a specific emotion is sent to a fully connected layer. Subsequently, a softmax layer predict the probability distributions of the emotion clause and the cause clause :
  \begin{eqnarray}
  \hat{y}_{i}^{e}=\operatorname{softmax}\left(W_{\mathrm{e}}^{\mathrm{T}} h_{i}^{\mathrm{e}}+b_{\mathrm{e}}\right) 
  \label{g7}
  \end{eqnarray}
  \begin{eqnarray}
  \hat{y}_{i}^{c}=\operatorname{softmax}\left(W_{\mathrm{c}}^{\mathrm{T}} h_{i}^{\mathrm{c}}+b_{\mathrm{c}}\right)
  \label{g8}
  \end{eqnarray}
  
  where $ W_e\in R^{2\times2d_h} $ and $ b_e\in R^2 $ are learnable parameters.

  Finally, we obtain cause-specific clause representations $ \left\{h_i^c\right\}_{i=1}^N $ , probability distributions over cause clauses $ \left\{{\hat{y}}_i^c\right\}_{i=1}^N $ , and updated cause label embeddings from the cause prediction module $ \left\{u_i^c\right\}_{i=1}^N $ . As well as the cause-specific clause representations $ \left\{h_i^e\right\}_{i=1}^N $ , probability distributions of the cause clauses $ \left\{{\hat{y}}_i^e\right\}_{i=1}^N $ , and updated cause label embeddings from the emotion prediction module $ \left\{u_i^e\right\}_{i=1}^N $ .

\subsection{Emotion cause pair prediction module}\label{subsec3}
\subsubsection{Knowledge graph filtering}\label{subsubsec3}

\begin{figure}[t]
\centering
\includegraphics[height=3cm,width=7.5cm]{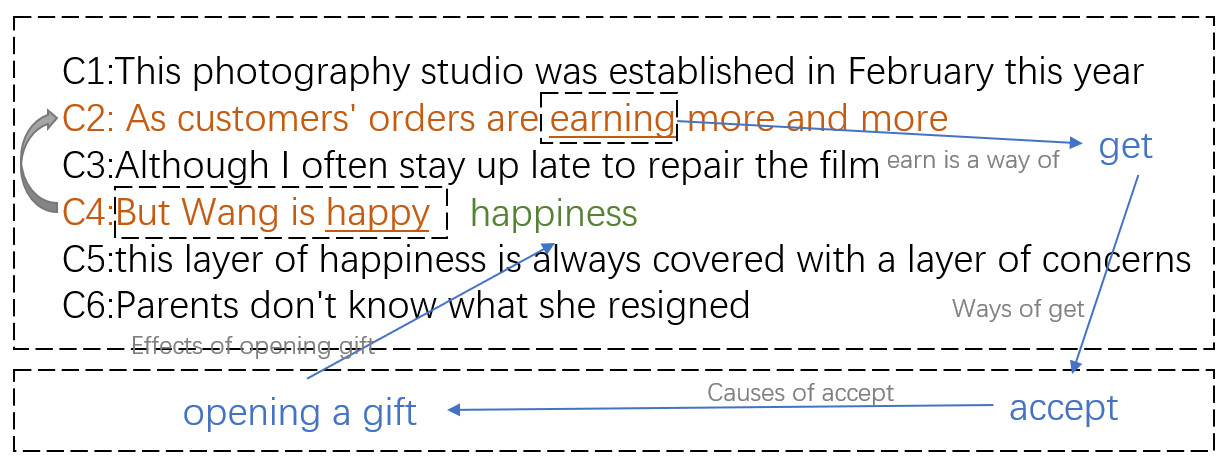}
\caption{Examples of knowledge graph filtering.}
\label{Figure4}
\end{figure}

Fig.\ref{Figure4} shows an example. The 4th clause is annotated as an emotion clause, and the emotion class label is "happiness". For the keyword "earn", we show example paths extracted from ConceptNet, each linking the word "earn" with "happy". One of the paths is "earn—get—accept—opening a gift—happiness—happy."

In the emotion-cause pair prediction module, contextual semantic information can be fully combined to learn the emotion-cause pair representation based on the relative position of the data and the knowledge graph. It can better capture the association between clause pairs, thereby improving the performance of emotion-cause pair extraction tasks.

The essence of external knowledge ConceptNet is a triplet containing start node, relation label, and end node. ConceptNet is a common-sense knowledge graph, which uses nodes to represent entities and the relationship between entities to represent entities. Considering the influence of causality on the relationship between candidate clauses and emotion clauses, We propose a causal path to extract words from candidate clauses and emotion clauses to label emotion words or emotion labels. Specifically, for the candidate clauses, We first use the Chinese word segmentation tool jieba word segmentation, and then extracts the top three keywords ranked by the TextRank algorithm. Each key-word of the candidate clause is regarded as a headentity $e_{h}$, and the emotion word or the emotion label of the emotion clause is used as the tail entity $e_{t}$. Igraph is used to conduct a depth-first search on ConceptNet, and it is determined to be the starting point $e_{h} e_{t}$ and the end point of the trajectory, and only keep the path containing less than 3 intermediate entities. The cause is that the shorter the path, the easier it is to give reliable causing evidence. This is due to the shorter average length of clauses, with many keywords existing within only one clause.

Specifically, if the absolute distance between clauses $c_{i}$ and clauses $c_{j}$ is less than or equal to a specific positive value ${W}$, We regard them as training samples for ECPE. Therefore, We construct a training set for ECPE $\mathcal{P}=\left\{\left(c_{i}, c_{j}\right)|| j-i \mid \leq\right.$ $W, i, j=1,2, \ldots, N\}$. In order to obtain a fine set of imbalanced test samples, for the $|j-i|>W$ imbalanced samples of, first use jieba to segment the emotion clause of a clause, and then use TextRank to extract the top three keywords. After that, each word in the candidate clause is regarded as the head entity, and the top three keywords of the tail entity are searched in ConceptNet. Use Igraph to search for the shortest path. If it can be reached within three hops, it means that there is a strong dependency between the candidate clause and the emotion clause. Filter out important imbalanced samples, and finally add them to the collection and use external knowledge to keyword in the $\mathcal{P}$.document Filter to get fine-grained semantically enhanced clause representations. In this way, delivering samples of fine-grained semantic features is beneficial to highlight the underlying causal features implicit in clause representations. The feature vectors for filtering positional imbalance test samples are initialized by the word embedding vectors published by Xia \cite{2_Xia2019semi}. For clause nodes, the context-aware corresponding clause representations generated by the clause-level encoder are used for initialization.

\subsubsection{Transformer layer}\label{subsubsec3}
  The transformer block is used to simulate the semantic interaction process between clauses and keywords, so as to better utilize the fine-grained semantic features contained in keywords and promote the representation learning of clauses. Consider that for a particular clause, different keywords have different importance. It intuitively realizes the fine-grained semantic connection between distant clauses, and helps to extract Emotion-cause pairs from distant clauses.

It can be seen from Fig. \ref{Figure2} that the distance between 98.15\% of the emotions and their corresponding causes is less than or equal to 3 . Therefore, We use a Transformer layer in the upper layer to encode the adjacent information and further enhance the embedding vector of each clause. Specifically, for the ith clause, emotion-cause pairs from indicator graph filtering are $v_{i}^{p}$ passed to the Transformer layer to generate neighborhood-encoded embedding vectors $h_{i}^{p}$. Then, we will $h_{i}^{p}$ be passed to have a softmax activation function to predict the uniform label of the $\hat{y}_{i j}$ ith clause.

\begin{eqnarray}
  h_{i}^{p}=\left[h_{i}^{c}, h_{i}^{e}\right] 
  \label{g9}
  \end{eqnarray}
  \begin{eqnarray}
  v_{i}^{p}= Transformer \left(h_{i}^{p}\right) 
  \label{g10}
  \end{eqnarray}

  Clauses in a document do not stand alone. Helping to grasp contextual clues can help the paper to better understand the current clause. Therefore, our standard encoder using The Transformer consists of N identical layers, with each layer having two layers. The first layer is a multi-head self-attention layer, and the second layer is a fully connected feedforward network.
  
\subsubsection{Emotion-cause pair prediction}\label{subsubsec3}

  The purpose of the emotion-cause pair prediction module is to combine the clauses into candidate pairs, and extract the emotion-cause pairs from the candidate clause pairs. Specifically, if the absolute distance between clause $c_ {i}$ and clause $c_ {j}$ is less than or equal to a positive value of $W$, we consider it as the training sample for ECPE. Therefore, we constructed a training set $P$ for ECPE.
\begin{figure}[t]
\centering
\includegraphics[height=4cm,width=7.5cm]{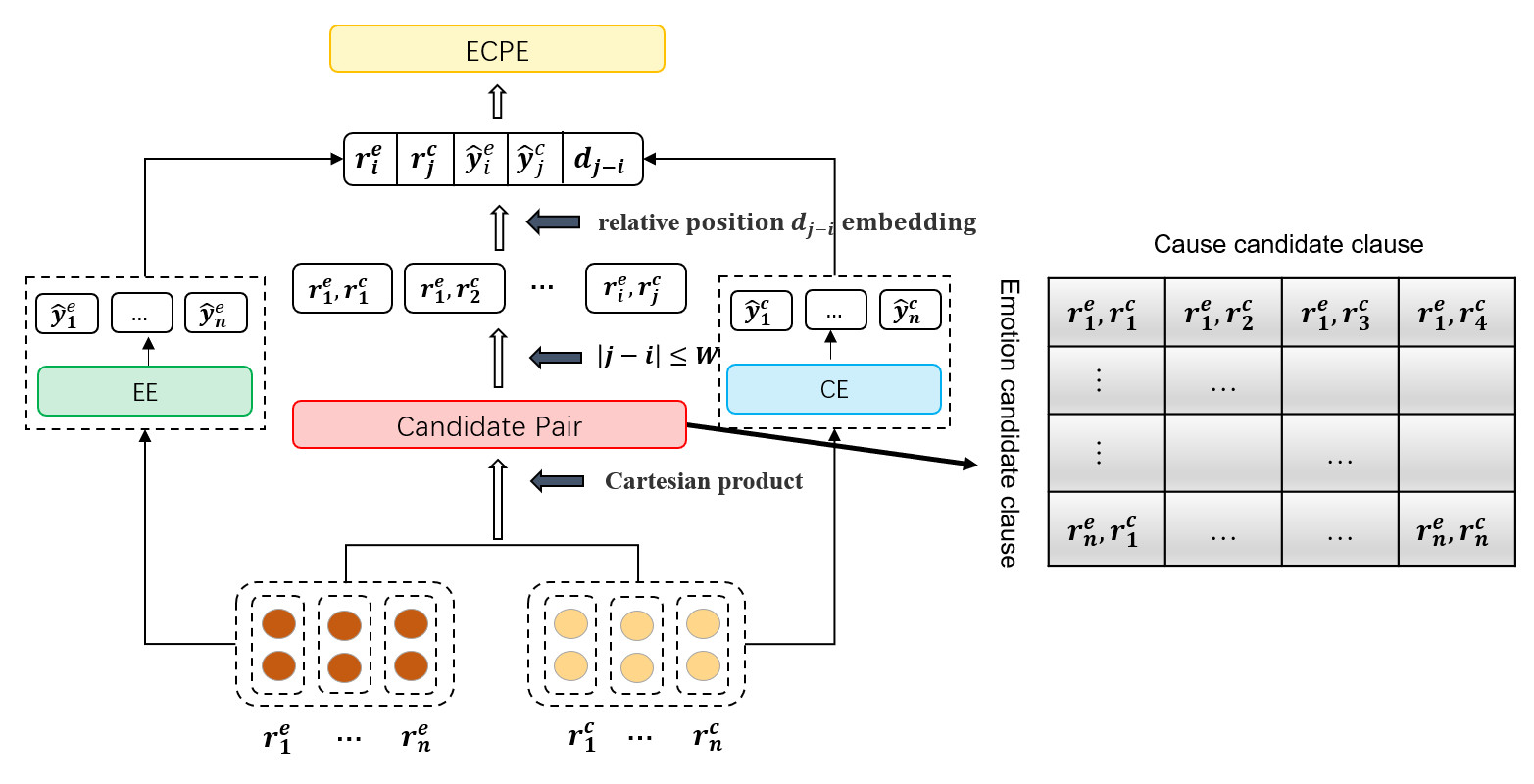}
\caption{Emotion cause pair prediction.}
\label{Figure5}
\end{figure}

  As shown in Fig.\ref{Figure5} , for each candidate clause pair $\left(c_{i}, c_{j}\right) \in P$, we construct its representation by concatenating two vectors, namely the clause of clause $c_{i}$ and the clause of clause $c_{j}$. The embed-ding constitutes the emotion-cause pair $v_{i}^{p}$, and their relative position embedding $\boldsymbol{d}_{j-i}^{p} \in R^{d_{r}}$, en-coding the distance between clauses $c_{i}$ and $c_{j}$. Then, we use a fully connected layer followed by a ReLU function to obtain a task-specific pair representation $\widehat{\boldsymbol{p}}_{i j}$ for ECPE.

  For a candidate clause pair $\left(c_{i}^{e}, c_{j}^{c}\right) \in P$, we represent it as $v_{i}^{p}=$ $\left[h_{i}^{c}, h_{i}^{e}, u_{i}^{c}, u_{i}^{e}\right]$ a fully connected layer with softmax activation function. Consider the role of distance in-formation between pairs of Emotion-causes. Concatenate candidate pairs with their distance embeddings. Predict its label:
  
\begin{eqnarray}
  \hat{p}_{i j}=\operatorname{Re} L U\left(W_{p}^{\top}\left[v_{i}^{p}, d_{j-i}^{p}\right]+b_{p}\right) 
  \label{g11}
  \end{eqnarray}

  where $\mathrm{W}_{p}$ and $\mathrm{b}_{p}$ are trainable parameters.

Then, the pair representations are $p_{i j}$ fed into a softmax classifier to obtain the probability that the candidate pair representations $\left(c_{i}^{e}, c_{j}^{c}\right)$ are emotion-cause pairs $\hat{y}_{i j}$ :

  \begin{eqnarray}
  {\hat{y}}_{ij}=softmax\left({\hat{W}}_pp_{ij}+{\hat{b}}_p\right)\ 
  \label{g12}
  \end{eqnarray}
  
During model training, we use two cross-entropy loss functions $\mathcal{L}_{\text {emo }}$ and sum $\mathcal{L}_{\text {cau}}$ to supervise the clause representation learning in the clause-level encoder, and use the cross- entropy loss function $\mathcal{L}_{\text {pair }}$ to supervise the final emotion-cause pair prediction. The loss functions of the three tasks are calculated as follows:
  \begin{eqnarray}
  \mathcal{L}_{e m o}=-\sum\left(y_{e}^{k} \log \left(\hat{y}_{e}^{k}\right)+\left(1-y_{e}^{k}\right) \log \left(1-\hat{y}_{e}^{k}\right)\right) 
  \label{g13}
  \end{eqnarray}
  \begin{eqnarray}
  {L}_{\text {cau }}=-\sum\left(y_{c}^{k} \log \left(\hat{y}_{c}^{k}\right)+\left(1-y_{c}^{k}\right) \log \left(1-\hat{y}_{c}^{k}\right)\right) 
  \label{g14}
  \end{eqnarray}
  \begin{eqnarray}
  {L}_{\text {pair }}=-\sum_{\forall\left(c_{i}, c_{j}\right) \in \mathcal{P}}\left(y_{i j} \log \left(\hat{y}_{i j}\right)+\left(1-y_{i j}\right) \log \left(1-\hat{y}_{i j}\right)\right) 
  \label{g15}
  \end{eqnarray}

  As a joint task, the loss function $\mathcal{L}$ formula is as follows:

  \begin{eqnarray}
  \mathcal{L}={\alpha\mathcal{L}}_{\mathrm{pair}}+{\beta\mathcal{L}}_{\mathrm{emo}}+\gamma\mathcal{L}_{\mathrm{cau}}+\lambda\parallel\theta\parallel^2 
  \label{g16}
  \end{eqnarray}

  where $\alpha, \beta$ and $\gamma$ are hyperparameters and $\lambda$ are the coefficients of the $\mathrm{L} 2$ regularization loss $\|$ $\theta \|^{2}$.

  \subsection{Position-aware interactive encoding module}\label{subsec4}

  In order to better solve the problem of sample imbalanced distribution of emotion clauses and cause clauses, We replace the original word coding layer with a position-aware coding layer. First, the sentiment lexicon is introduced in the text encoding layer to enhance the representation of the clauses, so that the sentences contain rich emotion information and contextual semantic information when encoding. Then, a position-aware interactive module (PAIM) is introduced in the shared coding layer to capture the position bias to alleviate the problem of sample Imbalanced distribution of emotion clauses and cause clauses. And fully combine the predictions obtained in the emotion extraction and emotion cause extraction modules to learn the emotion cause pair representation based on the relative position of the data, and better capture the relation-ship between the clause pairs.

   \subsubsection{Lexicon enhanced document encoding}\label{subsec1}

   We feed the entire document into the pretrained BERT\cite{24_Devlin2019bert}. To denote a single clause, we insert a [CLS] token at the beginning of each clause and a [SEP] token after each clause. We also use interval separator embeddings to distinguish between multiple clauses in a document. For clauses $c_{i}$, embed or label $E_{B}$ respectively depending on whether $E_{A} $ is odd or even. For example, for the text $\left[c_{1}, c_{2}, c_{3}\right.$,$\left.c_{4}, c_{5}\right]$, the text is embedded respectively $\left[E_{A}, E_{B}, E_{A}, E_{B}, E_{A}\right]$

   In addition, We combine emotion lexicon emotion vocabulary with sentences.We first use the emotion lexicon to retrieve vocabulary, combine related vocabulary with sentences, and then insert [CLS] at the beginning of the clause and insert tags at the end [SEP]. The output $h_{i}$ is used to represent clauses $c_{i}$, and the representation of the entire document is denoted as $H=\left\{h_{1}, h_{2}, \ldots, h_{n}\right\}$.

   \subsubsection{Position-aware interaction module}\label{subsec2}
   
\begin{figure}[t]
\centering
\includegraphics[height=5.6cm,width=7.5cm]{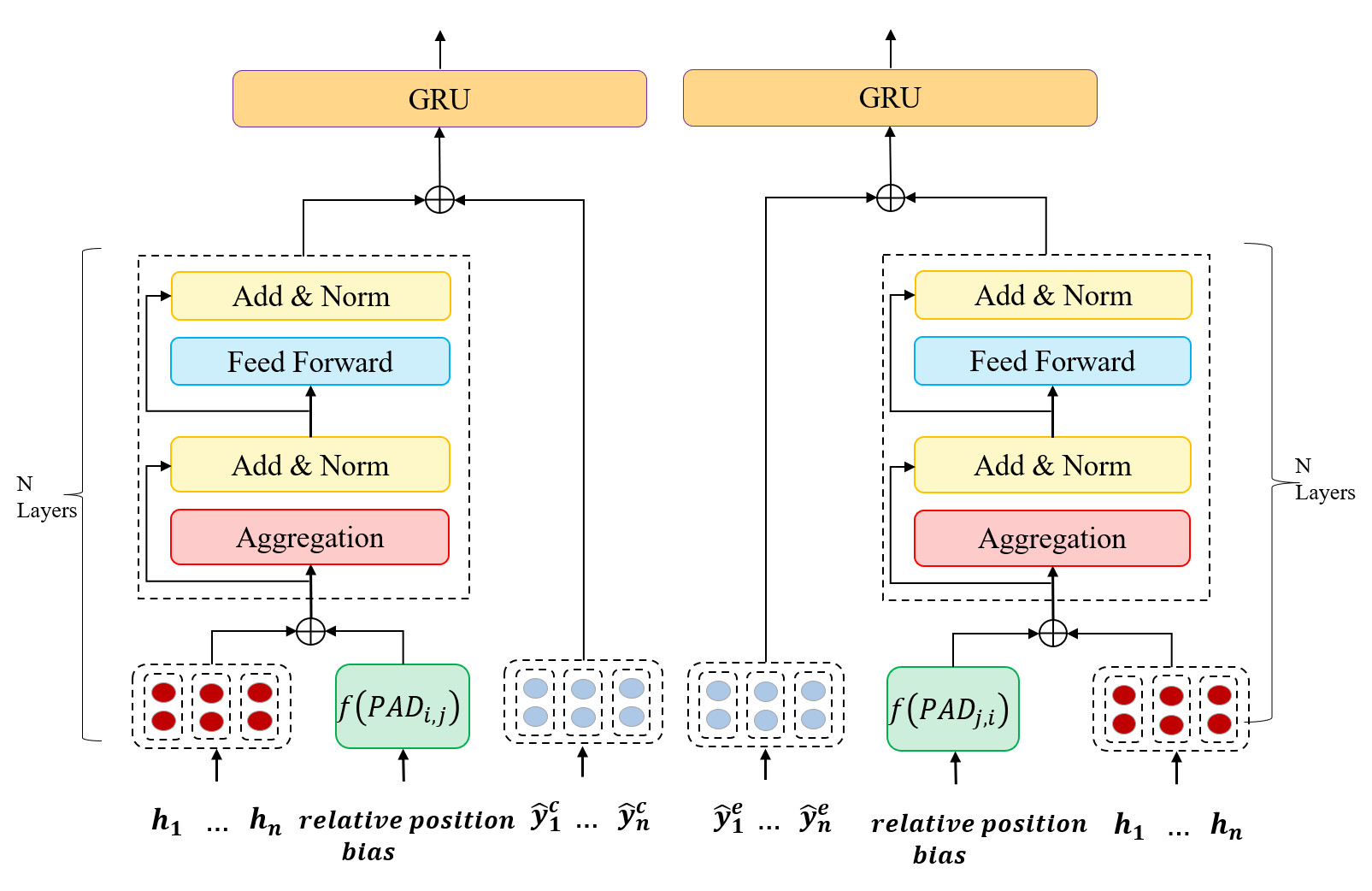}
\caption{Position-aware interaction module.}
\label{Figure6}
\end{figure}

   As shown in Fig.\ref{Figure6}, in order to capture the positional reciprocity between emotion and cause to solve the positional deviation problem in the context modeling stage, we design a perceptual interaction module PAIM, which includes feedforward network, self-attention and layer regression A unified Transformer encoder architecture. Furthermore, we combine self-attention with a trainable relative positional bias term as its core component to model positional bias.

   Since the clauses in the document are interrelated, mastering contextual clues can help the model better understand specific clauses. Therefore, we deploy the Transformer module to fuse the relevant information of other clauses with specific clauses to generate clause representations.

A multihead self-attention layer with positional bias was added. We adopt multi-head attention. For each sentence $c_{i}$, we first feed its sentence representation $h_{i} \in R^{d}$ into three different fully connected layers to generate query, key, value, expressed as $q_{i}, k_{i}$ sum $v_{i}$, as follows:
\begin{eqnarray}
  q_i=W_Qh_i 
  \label{g17}
  \end{eqnarray}
  \begin{eqnarray}
  k_i=W_Kh_i 
  \label{g18}
  \end{eqnarray}
  \begin{eqnarray}
  v_i=W_Vh_i 
  \label{g19}
  \end{eqnarray}
  \begin{eqnarray}
  s_{ij}=\frac{q_ik_j}{\sqrt d} 
  \label{g20}
  \end{eqnarray}

Among them, $W_{Q}, W_{K}, W_{V} \in R^{d \times d}$. The attention score $s_{i j}$ is the same as the standard self-attention mechanism, calculated by query, key, and value.

A trainable bias term is used to model the positional bias, which is then aggregated with the standard self-attention score to obtain the final self-attention representation.

The distance between the $|i-j|$ input emotion clause and the cause clause is determined by the PAD(Position -aware Distance) formula, indicating the absolute value distance between the emotion word and the cause word, and $m$ indicating the length of the emotion word or the cause word.
\begin{eqnarray}
  {PAD}_{i,j}=\left|i-j\right|-\left\lfloor\frac{m}{2}\right\rfloor 
  \label{g21}
  \end{eqnarray}
  \begin{eqnarray}
  e_{ij}=s_{ij}+f\left({PAD}_{i,j}\right) 
  \label{g22}
  \end{eqnarray}
  \begin{eqnarray}
  \ f\left(i-j\right)=\left\{\begin{matrix}g\left({PAD}_{i,j}\right),&\mathrm{\ if\ }\left|i-j\right|\le W\\b,&\mathrm{\ otherwise\ }\\\end{matrix}\right. 
  \label{g23}
  \end{eqnarray}

$g(i-j)$ is a bias function that limits $W$ the relative distance within the window, and further maps it to represent the corresponding position bias.

Obtaining the adjusted attention scores, $e_{i j}$ the value vectors are aggregated after the softmax operation, and then input into the encoding block:
\begin{eqnarray}
  z_{i}=\sum_{j} \operatorname{Softmax}\left(e_{i j}\right) v_{j}
  \label{g24}
  \end{eqnarray}

   The attention sublayer is followed by a fully connected feedforward network layer. We add a residual connection at the output of each sublayer to assist Transformer training, and then perform layer normalization. Finally, the Transformer's encoder outputs a set of sentence embeddings, denoted as $R=$ $\left[r_{1}, r_{2}, \ldots, r_{i}\right]$

Intuitively, different encoders can capture different types of features. Therefore, we use two clause-level position-aware modules $PAIM-E$ and obtain $PAIM-C$ emotion representation $R^{e}$ and cause representation $R^{c}$ for each clause respectively $c_{i}$.
\begin{eqnarray}
  R^e=PAIM^e (H)
  \end{eqnarray}
  \begin{eqnarray}
  R^c=PAIM^c\left(H\right) 
  \label{g25}
  \end{eqnarray}

The input of emotion clause encoding consists of the following two parts: clause representation from context encoding module $\left\{\boldsymbol{r}_{i}\right\}_{i=1}^{n}$, cause label embedding from cause prediction task $\left\{\widehat{\boldsymbol{y}}_{i}^{c}\right\}_{i=1}^{n}$, concatenate the two parts, and input to PAIM module to encode specific emotion context at clause level information
\begin{eqnarray}
  r_i^e=\left[r_i;{\hat{y}}_i^c\right] 
  \label{g27}
  \end{eqnarray}

\section{Experiment and analysis}
\subsection{Experimental dataset and evaluation metrics}\label{sec4}

We choose the intersection of HowNet and ANTUSD as the sentiment lexicon. We evaluate the model on the ECPE benchmark dataset. This dataset was constructed by Xia \cite{2_Xia2019semi}based on the Chinese Emotion-causes Corpus. Among them, the number of documents is 1945, and the statistical data of the specific data set are shown in Table \ref{table1}. 90\% of the data is selected for training, and 10\% of the data is used for testing. In order to obtain reliable results, each experiment was repeated 10 times and the average value was selected as the result.

\begin{table}[h]
\centering
\caption{Chinese dataset statistics.}
\label{table1}
\begin{tabular}{m{6cm}m{1cm}}
\hline 
Total number of documents & 1945 \\
\hline 
Number of documents with 1 pair & 1746 \\Number of documents with 2 pairs & 177 \\The number of documents with $\geq 3$ pairs & 22 \\Average number of clauses per document & 14.77 \\Average number of pairs per document & 1.11 \\Maximum number of clauses per document & 73 \\
\hline
\end{tabular}
\end{table}

\subsection{Experiment settings}\label{sec2}

We use 200-dimensional word embeddings to initialize word vectors, using the word $2\mathrm{vec}$ toolkit to pretrain word embeddings on a corpus of 1.1 million microblogs. In addition, We set the relative position embedding to 50 and the hyperparameter $W$ to 3. Dropout is applied to the word embedding layer with probability 0.5 . We use the Adam optimizer for training. Batch size set to 4, learning rate initialized to $2\mathrm {e}-5 $. The  coefficient of $L_ {2}$ regularization $\lambda $ is set to $1{e}-5 $ We have adopted BERT for the sentence encoder in our work. The number of dimensions indicated by the clause is set to 200. In Transformer, the number of layers of the encoder module is set to 2, and the dimension of the hidden state is set to 400. The query, key, and value vectors are all set to 200, and the relative distance is set to 4. Set the parameter set $\eta $to 0.5. In the experiment, we divided the data into 10 parts and used Xia \cite {2_Xia2019semi}10-fold Cross validation for evaluation.  We use precision $\mathrm{P}$, recall $\mathrm{R}$ and $\mathrm{f} 1$ score as evaluation metrics:
\begin{eqnarray}
  P=\frac{\sum\mathrm{\ correct_pairs\ }}{\sum\mathrm{\ predict_pairs\ }} 
  \label{g28}
  \end{eqnarray}
  \begin{eqnarray}
  R=\frac{\sum\mathrm{\ correct_pairs\ }}{\sum\mathrm{\ annotated_pairs\ }} 
  \label{g29}
  \end{eqnarray}
  \begin{eqnarray}
  F1=\frac{2\ast P\ast R}{P+R} 
  \label{g30}
  \end{eqnarray}

  \subsection{Comparison methods}\label{sec3}

  Existing models take a two-stage or end-to-end approach, and the method is compared with the following methods.
  
(1) Two-stage methods:

\begin{itemize}
\item Indep \cite{2_Xia2019semi}: It is a two-stage framework. First, the clauses are encoded with BiLSTM, and then BiLSTM is used to model the emotion and cause clauses respectively to extract emotion clauses and cause clauses, and then filter emotion-cause pairs.
\item Inter-CE \cite{2_Xia2019semi},Inter-EC \cite{2_Xia2019semi}: both are variants of Indep. Inter-CE leverages predictions from cause extraction to help emotion extraction.Inter-EC leverages predictions from emotion extraction to help cause extraction.
\item MAM-SD \cite{17_Yu2021Mutually}:It is a joint model of multi task interaction that can simulate the interaction between emotion extraction and reason extraction.
\end{itemize}

(2) End-to-end methods:

  \begin{itemize}
\item MTNECP \cite{15_Wu2020multi}: It is a Joint learning framework. It shares three tasks and exploits location information for different tasks.
\item TDGC\cite{23_Song2021An}: It constructs directed graphs with labeled edges to extract emotions and their corresponding causes.
\item ECPE-2D \cite{11_Ding2020ECPE-2D}: Based on multi-task learning, ECPE-2D uses multi-layer transformers to further capture the connections between clauses.
\item Inter-EC+WC \cite{11_Ding2020ECPE-2D}, Inter-EC+CR \cite{11_Ding2020ECPE-2D}: they are multi-task model based on ECPE-2D. Inter-EC+WC is a window-constrained 2D transformer used to build the interaction between different emotion-cause pairs.
Inter-EC+CR is a two-way interaction 2D transformers used to build the interaction between different emotion-cause pairs.
\item E2EECPE \cite{16_Song2021An}: E2EECPE is a typical multi-task learning method for the ECPE task. It treats emotion-cause pair extraction as a link prediction task, and uses auxiliary emotion extraction and cause extraction to facilitate pair extraction.
\item RankCP \cite{9_Hu2014FSS-GCN}: It leverages the correlation between clauses and kernel-based relative positional embeddings for efficient ranking.
\item RSN \cite{18_Tang2019Joint}: It is a recurrent synchronous network adopted to jointly extract emotions, causes, and emotion-cause pairs.
\item ECPE-MLL \cite{12_Ding2020End-to-end}: Based on sliding window multi-label learning, and integrates two task extractions with emotion and cause as pivots to obtain the final result in the prediction stage.
\item UTOS \cite{25_Cheng2021Unified}: It designs a unified labeling scheme, redefines the ECPE task as a sequence labeling problem, and proposes a unified object-oriented sequence to sequence model to solve the labeling problem.
\item MTST-ECPE \cite{26_Chen2022Recurrent}: It proposes a multi-task sequence labeling framework that can simultaneously extract emotions with associated causes by encoding the distance of emotions into a new labeling scheme.
\end{itemize}

\begin{table*}[h]
\centering
\caption{Comparison of models based on LSTM and GRU.}
\label{table2}
\begin{tabular}{m{3cm}m{0.7cm}m{0.7cm}m{0.7cm}m{0.7cm}m{0.7cm}m{0.7cm}m{0.7cm}m{0.7cm}m{0.7cm}}
\hline
\multirow{1}*{method} &
\multicolumn{3}{c}{Emotion Extraction} &
\multicolumn{3}{c}{Cause Extraction} &
\multicolumn{3}{c}{Emotion-cause Pair Ext}\\
\cmidrule{2-4} \cmidrule{5-7} \cmidrule{8-10}
&P &R &F1 & P & R & F1 & P & R & F1 \\
\hline
Indep &83.75	&80.71	&80.71	&69.02	&56.73	&62.05	&68.32	&50.82	&58.18 \\

Inter-CE &84.94	&81.22	&81.22	&68.09	&56.34	&61.51	&69.02	&51.35	&59.01  \\

Inter-EC &83.64	&81.07	&82.30	&70.41	&60.83	&65.07	&67.21	&57.05	&61.28\\

MAM-SD &85.54	&81.41	&83.39	&72.02	&63.75	&67.51	&69.63	&57.99	&63.20   \\

MTNECP &86.62	&83.93	&85.20	&74.00	&63.78	&68.44	&68.28	&58.94	&63.21  \\

TDGC &80.80	&84.39	&82.56	&67.42	&65.34	&66.36	&65.15	&63.54	&64.34  \\

Inter-EC+WC &85.11	&82.37	&83.65	&71.33	&62.85	&66.72	&71.18	&59.84	&64.94  \\

Inter-EC+CR &85.12	&82.20	&83.58	&72.72	&62.98	&67.38	&69.60	&61.18	&64.96  \\

E2EECPE &85.95	&79.15	&82.38	&70.62	&60.30	&65.03	&64.78	&61.05	&62.80  \\

RankCP &87.03	&84.06	&85.48	&69.27	&67.43	&68.24	&66.98	&65.46	&66.10  \\

RSN &86.88	&\textbf{87.43}	&87.07	&73.62	&65.54	&69.26	&72.15	&63.77	&67.62  \\

MM-ECPE &\textbf{88.06}	&87.22	&\textbf{87.64}	&\textbf{76.82}	&\textbf{69.53}	&\textbf{72.99}	&\textbf{73.40}	&\textbf{66.06}	&\textbf{69.54}  \\
\hline
\end{tabular}
\end{table*}

\begin{table*}[h]
\centering
\caption{ Comparison of models based on BERT.}
\label{table3}
\begin{tabular}{m{3cm}m{0.7cm}m{0.7cm}m{0.7cm}m{0.7cm}m{0.7cm}m{0.7cm}m{0.7cm}m{0.7cm}m{0.7cm}}
\hline
\multirow{1}*{method} &
\multicolumn{3}{c}{Emotion Extraction} &
\multicolumn{3}{c}{Cause Extraction} &
\multicolumn{3}{c}{Emotion-cause Pair Ext}\\
\cmidrule{2-4} \cmidrule{5-7} \cmidrule{8-10}
&P &R &F1 & P & R & F1 & P & R & F1 \\
\hline
TDGC(BERT) &87.16	&82.44	&84.74	&75.62	&64.71	&69.74	&73.74	&63.07	&67.99 \\

Inter-EC+WC(BERT) &86.27	&92.21	&89.10	&73.36	&69.34	&71.23	&72.92	&65.44	&68.89  \\

Inter-EC+CR(BERT) &85.48	&\textbf{92.44}	&88.78	&72.72	&69.27	&70.87	&69.35	&67.85	&68.37\\

RankCP (BERT) &91.23	&89.99	&90.57	&74.61	&\textbf{77.88}	&76.15	&71.19	&76.30	&73.60   \\

ECPE-MLL (BERT) &86.08	&91.91	&88.86	&73.82	&79.12	&76.30	&77.00	&72.35	&74.52  \\

UTOS(BERT) &88.15	&83.21	&85.56	&76.71	&73.20	&74.71	&73.89	&70.62	&72.03  \\

MTST-ECPE(BERT) &85.83	&80.94	&83.21	&77.64	&72.36	&74.77	&75.78	&70.51	&72.91  \\

MM-ECPE(BERT) &\textbf{92.91}	&90.25	&\textbf{91.56}	&\textbf{79.06}	&77.36	&\textbf{78.20}	&\textbf{77.48}	&\textbf{76.35}	&\textbf{76.91}  \\
\hline
\end{tabular}
\end{table*}

\subsection{Evaluation and comparison}\label{sec5}

  \textbf{1. MM-ECPE model:}

  Table \ref{table2} are models based on LSTM or GRU.

  (1) On the whole, the end-to-end model is better than the two-stage model. The end-to-end joint task takes into account the interaction among multiple tasks to reduce error propagation, demonstrating the effectiveness of MM-ECPE. Secondly, compared with the RSN model, the F1 value of the model  is increased by 1.92\%. For the MAM-SD model, our model improves the P, R and F1 scores by 3.77\%, 8.07\%, and 6.34\%, respectively. 

  (2) In addition, it can be observed in the table that RSN (BERT) extracts more positive cases than RANKCP (BERT) and the model are less. RSN (BERT) jointly executes EC, EE and ECPE, making full use of the relationship between each subtask. The model surpasses it a lot in R, because We screen out the main Emotion-cause pairs through the absolute value distance before performing the ECPE task, and use the knowledge graph to screen the distant Emotion-cause pairs to construct candidate cause pairs. On the one hand, this helps the model to find pairs with emotion-cause relationships, and on the other hand, it helps to alleviate the problem of sample location imbalance. Due to the multi-task model and multi-level sharing of sub-sentence representations output by modules, each task generates a task-specific representation, and our model also performs well on individual sub-tasks.

\textbf{2. MM-ECPE (BERT) model:}

The results in Table \ref{table3} are the results of the BERT-based method. We implement the BERT - based method MM-ECPE (BERT), which facilitates comparison with models based on the BERT method. The MM-ECPE (BERT) model replaces the original coding layer with a position-aware interactive coding layer, and the other parts are the same as the GRU-based MM-ECPE.

(1) Compared with other models, the MM-ECPE (BERT) model achieves better performance of F1 score on ECPE. In addition, the P and R scores outperform most models, proving the effectiveness of MM-ECPE (BERT). The MM-ECPE (BERT) is superior to other methods in the main task Emotion-cause and the extraction task, and in the two sub-tasks, the score is also among the best. In addition, RankCP and ECPE-2D perform worse on EE than on ECPE. At the same time, the methods that perform well on cause extraction generally perform better on Emotion-cause pair extraction, and the performance of each method on cause extraction is significantly lower than that of emotion extraction. We guess that the cause is that the cause extraction restricts the Emotion-cause to the extraction performance bottleneck. So, improving cause extraction is more beneficial to improving Emotion-cause extraction performance than improving emotion extraction.

(2) In multiple tasks, MM-ECPE (BERT) has achieved good results. Compared with the best models on the existing EE, CE and ECPE tasks, the method increases the F1 score by 0.99\%, 1.91\% and 2.39\%, respectively. The cause is that after the three tasks in MM-ECPE (BERT) interact with each other, EE sacrifices a small part of the performance to help ECPE and CE to obtain more obvious improvements. After combining the expression of emotion vocabulary to strengthen the clauses, the performance of the subtasks has been significantly improved. The results show that the pretraining BERT can effectively enhance the ability of clause-level contextual feature representation by using the vocabulary expansion clause representation of the sentiment lexicon. Compared with previous state-of-the-art methods, MM-ECPE (BERT) has a significant improvement on the F1 score in the ECPE task due to the use of emotion vocabulary. Since MM-ECPE(BERT) can effectively utilize the mutuality between the three sub-tasks, when the overall task performance is improved, MM-ECPE(BERT) can achieve better performance.

  (3) In terms of F1 performance, BERT-based methods RANKCP (BERT) and ECPE-MLL (BERT) are currently two better models. The model surpasses them a lot in terms of R value, because we screen out the main Emotion-cause pairs by adding position bias before performing the ECPE task helps alleviate the problem of imbalanced sample positions. Furthermore, due to the end-to-end model and the clause representations output by the multi-level shared modules, three tasks generate a task-specific representation, and our model also performs well on individual subtasks.

\subsection{Ablation study}\label{sec6}
\subsubsection{MM-ECPE ablation study}\label{subsubsec1}

In order to evaluate the impact of each component on the performance of the model, We delete each component in order to prove its effectiveness.

\textbf{MM-ECPE}: prototype model.
\textbf{MM-ECPE w/o KG}: Based on the model, the knowledge graph that introduces external knowledge is removed.
\textbf{MM-ECPE w/o inter}: Remove the interaction prediction information of emotion and cause on the basis of the model.
\textbf{MM-ECPE w/o pos}: Remove relative position information based on the model.
The final ablation experiments are shown in the table.

\begin{table}[h]
\centering
\caption{ MM-ECPE ablation experiments.}
\label{table4}
\begin{tabular}{m{3cm}m{0.8cm}m{0.8cm}m{0.8cm}}
\hline
\multirow{1}*{method} &
\multicolumn{3}{c}{Emotion-Cause Pair Ext} \\
\cmidrule{2-4} 
&P &R &F1\\
\hline
MM-ECPE	&\textbf{73.40}	&\textbf{66.06}	&\textbf{69.54}	\\

MM-ECPE w/o KG	&72.55	&65.20	&68.68	 \\

MM-ECPE w/o inter	&70.27	&65.90	&68.01	\\

MM-ECPE w/o pos	&71.65	&65.45	&68.41	\\
\hline
\end{tabular}
\end{table}

\begin{table}[h]
\centering
\caption{ MM-ECPE(BERT) ablation experiments.}
\label{table5}
\begin{tabular}{m{3.6cm}m{0.8cm}m{0.8cm}m{0.8cm}}
\hline
\multirow{1}*{method} &
\multicolumn{3}{c}{Emotion-Cause Pair Ext} \\
\cmidrule{2-4} 
&P &R &F1\\
\hline
MM-ECPE (BERT)	&\textbf{77.48}	&\textbf{76.35}	&\textbf{76.91}	\\

MM-ECPE(BERT) w/o lex	&73.05	&75.10	&74.06	 \\

MM-ECPE(BERT) w/o inter	&71.27	&75.70	&73.42	\\

MM-ECPE(BERT) w/o PAIM	&72.56	&75.45	&73.97	\\
\hline
\end{tabular}
\end{table}

It can be seen from Table \ref{table4} that 1) the F1 index of MM-ECPE w/o KG has decreased by 0.86\%, indicating that adding imbalanced samples screened by knowledge graph helps to improve the performance of the model. 2) Compared with the original model, the F1 index of MM-ECPE w/o inter dropped by 1.53\%, indicating the effectiveness of the multi-layer shared interaction layer, which is greatly improved compared with other single-layer interaction layer models. Positive effects of two auxiliary tasks, emotion prediction and cause prediction, on extraction of main task emotion and cause. The results of emotion prediction are helpful for cause extraction, and there is a mutual referential relationship between the two. 3) MM-ECPE w/o pos the 1.13\% decrease in the F1 index shows that the relative positional embedding between clause pairs does have a positive impact on the task.

\subsubsection{MM-ECPE (BERT) ablation study}\label{subsubsec2}
In order to evaluate the impact of each compo-nent on the performance of the model, We delete each component in order to prove its effectiveness.
MM-ECPE(BERT) w/o PAIM: Remove the position-aware interaction module PAIM on the basis of the model.
MM-ECPE(BERT) w/o lex: Remove the sen-timent lexicon based on the model.
MM-ECPE(BERT) w/o inter: On the basis of the model, the mutual information based on prior knowledge of the shared coding layer is removed.

The final ablation experiments are shown in Table \ref{table5}.

(1) Compared with the original model, the F1 index of MM-ECPE(BERT) w/o inter decreased by 3.44\%, indicating the effectiveness of the shared coding layer, which is greatly improved compared with other single-layer interactive layer models. The two subtasks emotion prediction and cause prediction have a positive impact on the main task emotion cause prediction. The results of emotion prediction are helpful for cause extraction, and there is a mutual referential relationship between the two.

(2) The F1 index of MM-ECPE(BERT) w/o PAIM dropped by 3.57\%, indicating that the position bias term has a positive impact on the task. By combining the position bias term of PAIM, each clause encodes emotion-specific representations Both focus on the previous clauses, and the final self-attention flow is consistent with the positional deviation between the emotion corresponding causes, enabling the model to focus on semantically consistent clauses around the target position, thereby eliminating noise. On the one hand, high-confidence clause information is exchanged between the two subtasks EE and CE, which helps the model find better optimization directions. On the other hand, the performance of the full model on ECPE is also better than the model using a single PAIM for clause representation, indicating that modeling the emotion clause and the cause clause with two PAIMs can help the model obtain more meaningful and specific emotion and specific cause representations further improve the performance of Emotion-cause pair extraction.

(3) The F1 index of MM-ECPE(BERT) w/o lex dropped by 1.97\%, indicating that the introduction of external knowledge through the sentiment lexicon to enhance the clause representation can help improve the performance of the model.

\subsection{Effectiveness of joint loss function}\label{sec6}

\begin{table}[h]
\centering
\caption{ Effectiveness of joint loss function.}
\label{table6}
\begin{tabular}{m{3cm}m{1cm}m{1cm}m{1cm}}
\hline
\multirow{1}*{Methods} &
\multicolumn{3}{c}{Emotion-Cause Pair Ext} \\
\cmidrule{2-4} 
&P &R &F1\\
\hline
MM-ECPE w/o Aux	&66.48	&64.22	&63.13\\
MM-ECPE	&\textbf{73.40}	&\textbf{66.06}	&\textbf{69.54}\\
\hline
\end{tabular}
\end{table}

In order to explore the impact of the auxiliary tasks of the multi-task learning model on the main task, We conduct ablation experiments on MM-ECPE. Specifically, this article removes the loss function and $\mathcal{L}_{\text {emo }}+\mathcal{L}_{\text {cau }}$ only uses it to train the model. This article calls it $\mathcal{L}_{\text {pair }}$ MM-ECPE w/o Aux. The P, R, and F 1 values of the ECPE task are used for evaluation, and the results are shown in Table \ref{table6}. We notice that the performance of MM-ECPE w/o Aux decreases compared with MM-ECPE, which shows that ECPE actually benefits from the joint learning of the three tasks.

F1 value of the emotion cause pair extraction task drops greatly. It is shown that two auxiliary tasks help to learn emotion features and causal features, thereby improving the performance of the main task.

\subsection{Influence of MM-ECPE Transformer layers}\label{sec7}
\subsubsection{MM-ECPE Transformer layer experiment}\label{subsubsec1}

\begin{table}[h]
\centering
\caption{ Effect of transformer layers.}
\label{table7}
\begin{tabular}{m{1cm}m{1cm}m{1cm}m{1cm}}
\hline
\multirow{1}*{layers} &
\multicolumn{3}{c}{Emotion-Cause Pair Ext} \\
\cmidrule{2-4} 
&P &R &F1\\
\hline
1	&71.62	&\textbf{66.26}	&68.84\\
2	&\textbf{73.40}	&66.06	&\textbf{69.54}\\
3	&71.20	&66.10	&68.56\\
4	&71.37	&65.05	&68.06\\
\hline
\end{tabular}
\end{table}

\begin{table}[h]
\centering
\caption{ Effect of the number of PAIM layers.}
\label{table8}
\begin{tabular}{m{1cm}m{1cm}m{1cm}m{1cm}}
\hline
\multirow{1}*{layers} &
\multicolumn{3}{c}{Emotion-Cause Pair Ext} \\
\cmidrule{2-4} 
&P &R &F1\\
\hline
1	&73.82	&75.26	&74.53\\
2	&\textbf{77.48}	&\textbf{76.35}	&\textbf{76.91}\\
3	&73.20	&75.09	&74.12\\
4	&73.37	&74.81	&74.08\\
\hline
\end{tabular}
\end{table}

Transformer affects the performance of Emotion-causes in the extraction process. Therefore, We conduct comparative experiments on different attention layers, and test the impact of different layers of Transformer modules on the performance of the model, and takes the performance of the model under different layers. The results show that when more Transformer layers and more parameters are used, the classification performance decreases. The test results are shown in Table \ref{table7}.

The model achieves the best performance when the number of Transformer layers is 2. Afterwards, as the number of layers increases, the parameters gradually increase, and the final performance continues to decline. When the number of Transformer layers is 2, in the MM-ECPE model, Emotion-causes achieve the best results in extraction performance. Experiments show that as the number of stacked layers increases, which in turn affects the performance of the model.

\subsubsection{Effectivness of MM-ECPE(BERT) PAIM layers}\label{subsubsec2}

Layers of PAIM will affect the performance when Emotion-causes are extracted. Therefore, this study conducts comparative experiments on models under different PAIM layers. In order to test the impact of different module layers on the performance of the model, the performance of the model under different layers was studied. The results show that when the number of layers is more or less, the classification effect decreases. The test results are shown in Table \ref{table8}.

It can be seen that when the number of layers is 2, the model achieves the best performance. Since then, with the increase of the number of layers, the parameters also increase, and the performance is get-ting worse. When the number of layers of PAIM module reaches 2, the performance of MM-ECPE (BERT) model in extracting Emotion-causes is the best.

\subsection{Experimental comparison of imbalanced samples}\label{sec8}

\begin{table}[h]
\centering
\caption{Experiments on imbalanced test$_{\text {imbalance}}$}
\label{table9}
\begin{tabular}{m{4cm}m{1cm}}
\hline 
Model & F1 \\
\hline 
Inter-EC	&34.28\\
RankCP	&39.17\\
ECPE-2D	&39.06\\
MM-ECPE	&\textbf{45.67}\\
MM-ECPE w/o KG	&40.48\\
MM-ECPE(BERT)	&\textbf{45.46}\\
MM-ECPE(BERT) w/o KG	&40.42\\
\hline
\end{tabular}
\end{table}

In addition, considering the problems caused by the positional bias, the original test set is divided, and the documents contain multiple Emotion-cause pairs and the positional imbalance samples test $_{\text {imbalance }}$ with the relative distance between clauses greater than 1 . And test the F1 score on the new test sample.

In order to evaluate the model's ability to deal with samples with imbalanced positions, We evaluate it on new test samples. It can be seen from Table \ref{table9} that the MM-ECPE model achieves the best results on the data imbalance test set. In addition, the model MM-ECPE w/o KG, which removes knowledge graph filtering, is 5.19\% lower than the original model, indicating that it can effectively alleviate the data imbalance. come the question.

As shown in Table \ref{table9}, in order to evaluate the ability of the model to deal with imbalanced samples, We select imbalanced samples whose relative distance between the emotion and the cause clause is greater than 1 as the test sample. Evaluation shows that MM-ECPE (BERT) model achieves the best results on data imbalanced samples. In addition, the model MM- ECPE (BERT) w/o PAIM with PAIM removed is 5.04\% lower than the original model, indicating that the PAIM module helps alleviate the problems caused by data imbalance.

\begin{table*}[h]
\centering
\caption{Case studies among different models.}
\label{table10}
\begin{tabular}{m{0.5cm}m{8.5cm}m{1cm}m{1cm}m{1cm}m{1.5cm}m{0cm}m{0cm}}
\hline
\multirow{1}*{ID} &
\multirow{1}*{Example} &
\multicolumn{3}{c}{Predicted results} &
\multicolumn{1}{c}{Ground Truth}\\
\cmidrule{3-5}
& & Inter-EC & RankCP  & MM-ECPE & \\
\hline
1 &...$\left(c_{1}\right)$ \textcolor{blue}{the student he once sponsored has graduated from university and has a job }$\left(c_{2}\right)$ \textcolor{red}{the heart of the old couple is relieved} $\left(c_{3}\right)$ now the children often call them or visit them $\left(c_{4}\right)$ ...
&	null & $c_3-c_2$ &	$c_3-c_2$ & $c_3-c_2$ \\
\hline
2 & ... suddenly a woman with a baby in her arms appeared at the door $\left(c_{3}\right)$ \textcolor{red}{with an anxious face} $\left(c_{4}\right)$ crying and shouted to Wang Xiaojun Master $\left(c_{5}\right)$ ... his body was still twitching $\left(c_{10}\right)$ \textcolor{blue}{he guessed that the child suddenly had a sudden illness} $\left(c_{11}\right)$
& $c_4-c_3$ & null 
& $c_4-c_{11}$ & $c_4-c_{11}$ \\

\hline
\end{tabular}
\end{table*}

\begin{table*}[h]
\centering
\caption{Case studies among different models.}
\label{table11}
\begin{tabular}{m{0.5cm}m{8.5cm}m{1cm}m{1cm}m{1cm}m{1.5cm}m{0cm}m{0cm}}
\hline
\multirow{1}*{ID} &
\multirow{1}*{Example} &
\multicolumn{3}{c}{Predicted results} &
\multicolumn{1}{c}{Ground Truth}\\
\cmidrule{3-6}
& & RankCP & — w/o PAIM & MM-ECPE (BERT) & \\
\hline
1 &...He always believed that the younger brother would suddenly open his eyes $\left(c_{4}\right)$ \textcolor{blue}{the younger brother had hope} $\left(c_{5}\right)$ and the parents would be\textcolor{red}{ happier} $\left(c_{6}\right)$ 
&	$\left(c_{6}, c_{5}\right)$ & $\left(c_{6}, c_{5}\right)$ &	$\left(c_{6}, c_{5}\right)$ & $\left(c_{6}, c_{5}\right)$ \\
\hline
2 & ...Compared with some time ago $\left(c_{2}\right) \mathrm{Mr}$. Sun's mental state was much better $\left(c_{3}\right)$\textcolor{blue}{Gradually walked out of the} \textcolor{red}{sadness} \textcolor{blue}{of losing his daughter} $\left(c_{4}\right)$ \textcolor{blue}{Facing so many people's concern} $\left(c_{5}\right)$ He felt I am very \textcolor{red}{happy} $\left(c_{6}\right)$ ... 
& $\left(c_{4}, c_{4}\right)$ & $\left(c_{4}, c_{4}\right)$ 
& $\left(c_{4}, c_{4}\right)$ $\left(c_{6}, c_{5}\right)$ & $\left(c_{4}, c_{4}\right)$ $\left(c_{6}, c_{5}\right)$ \\
\hline
3 & ... \textcolor{blue}{Although Tia 's birth made Sam's family very}\textcolor{red}{happy} $\left(c_{5}\right)$, Kelly, who has broken up with Sam's ex-girlfriend and just became a grand-mother, is a little \textcolor{red}{worried} $\left(c_{6}\right)$ ...all the time $\left(c_{10}\right)$ \textcolor{blue}{my worst nightmare was that Tia would repeat My mistake} $\left(c_{11}\right)$ ...
& $(c_5,c_5)$ & $(c_5,c_5)$ &	$(c_5,c_5)$ $(c_6,c_{11})$ & $(c_5,c_5)$ $(c_6,c_{11})$ \\
\hline
\end{tabular}
\end{table*}

\subsection{Influence of MM-ECPE (BERT) parameters {W}}\label{sec9}

\begin{figure}[t]
\centering
\includegraphics[height=3cm,width=6cm]{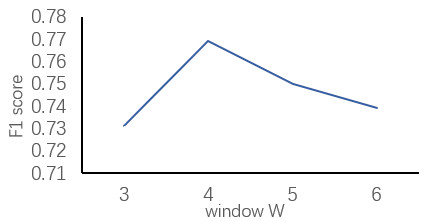}
\caption{The effect of parameter W.}
\label{Figure7}
\end{figure}

Instead of using all candidate pairs as the training set,We build the ECPE task training set and filtering candidate pairs $c_{i}^{e}, c_{j}^{c}$ whose relative distance $|\mathrm{j}-\mathrm{i}|$ between the emotion clause and the cause clause is less than or equal to the window $W$. In addition, we explore the effectiveness of training set construction methods through ablation studies.

The training set is constructed according to the hyperparameters, Wand the influence of different values of the hyperparameters on the model is further explored. The results are shown in Fig.\ref{Figure7}. When W to 4, our model achieves the best performance. This proves that our method of screening emotion-cause pairs by window size is effective, because it effectively alleviates the problem of label Imbalanced distribution of emotion clauses and cause clauses.

\subsection{Case study}\label{sec10}
\subsubsection{MM-ECPE case study}\label{sec1}

Table \ref{table10}  demonstrates the advantages of our multi-task end-to-end model over the two-stage framework model Inter-EC through two examples. In the first example, the Inter-EC model failed to extract emotion-cause pairs $c_{3}-c_{2}$. In contrast, our MM-ECPE model successfully identified this pair. In the second example, Inter-EC failed to extract the correct emotion-cause pair. By observing the results of Inter-EC, we find $c_{4}$ that the predicted clauses are emotion clauses and $c_{3}$ the predicted clauses are cause clauses. The filter ultimately fails to eliminate invalid pairs $c_{4}-c_{3}$. In contrast, our model is more reliable. The multi-task model uses the samples filtered by the knowledge graph. These circumstances make our model have higher accuracy performance, and make our model robust.

Compared with the higher performance model RankCP. For the first input example, include emotion-cause pairs $c_{3}-c_{2}$. The former is emotion, the latter is cause. As mentioned earlier, RankCP treats the predicted top-ranked pairs as emotion-cause pairs. For the remaining emotion-cause pairs, if the pair contains an emotion word in the external emotion lexicon, RankCP extracts it as an emotion-cause pair. It's a pity that there are no explicit emotion words in the first document. Therefore, RankCP only selects the top-ranked pair and cannot extract the correct emotion- cause pair. In contrast, our MM-ECPE successfully extracts emotion-cause pairs $c_{3}-c_{2}$. In the second document example, although the clause $c_{4}$ contains the explicit emotion word "anxiety", the position between this pair is too large for RankCP to recognize. In contrast, our method achieves better performance without these constraints.

\subsubsection{MM-ECPE(BERT) case study}\label{sec2}

Three examples are selected to analyze the performance difference between the model  and the model with better performance in the ECPE task and the ablation model . Table \ref{table11} shows the prediction results of different models under three examples.

is compared with the higher performance RankCP model. For the first example, both RankCP and our model can extract emotion-cause pairs well $\left(c_{6}, c_{5}\right)$. For the second example, two Emotion-cause pairs and $\left(c_{4}, c_{4}\right)$ sums are included $\left(c_{6}, c_{5}\right)$. RankCP ranks the predicted pairs, and the top-ranked pair is regarded as the Emotion-cause pair, and the correct Emotion-cause pair cannot be extracted. In contrast, our MM-ECPE (BERT) success-fully extracts two emotion cause pairs due to the introduction of external emotion vocabulary. In the third example, although the clause $c_{6}$ contains the clear emotion word "worried", the relative position of the distance from the cause clause is too large to be recognized by RankCP, and $c_{11}$ the MM-ECPE (BERT) of We will be enhanced according to the position It is extracted as the predicted result. This article can be found from the above examples. In contrast, our method achieves better performance.

Compare with our ablation model MM-ECPE(BERT) w/o PAIM. In the first and second examples, both our model and the ablation model correctly extract emotion-cause pairs $\left(c_{6}, c_{5}\right)$, and in the third example, the ablation model produces fewer emotion-cause pairs because the ablation model ignores the position-imbalanced samples yes $\left(c_{6}, c_{5}\right)$. In contrast, our full model successfully identifies all three examples. Therefore, our PAIM module alleviates the confusion problem when multiple pairs are in one document.

\section{Conclusion}
We propose a novel ECPE framework MM-ECPE. Although previous end-to-end methods involve many fields and methods, none of them consider how to understand other syntactic structures from one clause. Our MM-ECPE approach alleviates error propagation for two-stage tasks through an end-to-end recurrent model. A multi-level shared network is used to fully learn the mutual information between emotion and cause, and when generating emotion clause representations and cause clause representations, the mutual relationship between emotion clauses and cause clauses is fully considered. Filter the samples through the knowledge graph to solve the problem of imbalanced labels in the data set, and then learn the candidate Emotion-cause pair information through the transformer, and combine the position information, thereby improving the extraction performance. It employs multi-task learning to correctly extract different information and models the interactions among three tasks. 

Although the performance of our model has been improved, there are still several problems as follows. First, current research is mostly coarse-grained, at-tempting to design models to address fine-grained causes of emotion. This article will try to design a unified framework to extract emotions, and then extract their corresponding causes around the context of emotions. In addition, it is planned to develop more advanced models, such as heterogeneous graph neural networks, to combine with learned position information instead of refined position information. Finally, the problem of label imbalance should be solved with task-specific strategies such as contrastive learning. Future research will focus on these issues.

\section{Acknowledgments}
This work was supported by the National Natural
Science Foundation of China (grant No.62066022).

\bibliographystyle{unsrt}
\bibliography{bibliography}

\end{document}